\documentclass{Interspeech2024}

\usepackage{xurl}
\usepackage{dblfloatfix}
\usepackage{dirtytalk}
\usepackage{tablefootnote}
\usepackage{xcolor}
\usepackage{multirow}

\interspeechcameraready

\title{Automatic Classification of News Subjects in Broadcast News: Application to a Gender Bias Representation Analysis}

\name[affiliation={1}]{Valentin}{Pelloin}
\name[affiliation={2}]{Lena}{Dodson}
\name[affiliation={1}]{Émile}{Chapuis}
\name[affiliation={1}]{Nicolas}{Hervé}
\name[affiliation={1}]{David}{Doukhan}

\address{
  $^1$French National Institute of Audiovisual (INA), Paris, France\\
  $^2$French Digital Communication Regulatory Authority (ARCOM), Paris, France}
\email{\{vpelloin,echapuis,nherve,ddoukhan\}@ina.fr, lena.dodson@arcom.fr}


\begin{document}

\maketitle

\begin{abstract}
This paper introduces a computational framework designed to delineate gender distribution biases in topics covered by French TV and radio news.
We transcribe a dataset of 11.7k hours, broadcasted in 2023 on 21 French channels.
A Large Language Model (LLM) is used in few-shot conversation mode to obtain a topic classification on those transcriptions.
Using the generated LLM annotations, we explore the finetuning of a specialized smaller classification model, to reduce the computational cost. 
To evaluate the performances of these models, we construct and annotate a dataset of 804 dialogues. This dataset is made available free of charge for research purposes.
We show that women are notably underrepresented in subjects such as sports, politics and conflicts. Conversely, on topics such as weather, commercials and health, women have more speaking time than their overall average across all subjects. We also observe representations differences between private and public service channels. 
\end{abstract}

\section{Introduction}
    In April 2018, the European Parliament passed resolutions to improve gender equality in the media sector in the EU. One of its resolutions encourages \say{public and private media to mainstream gender equality in all their content}\footnote{\scriptsize\url{https://eur-lex.europa.eu/legal-content/EN/TXT/?uri=CELEX:52018IP0101}}.
    More generally, multiple studies have been conducted to monitor and promote equal gender representation in the media \cite{ComasdArgemir_2009,Walsh_Suiter_OConnor_2015,arcom24,macharia2020global}. 

    In France, since 2016, the Regulatory Authority for Audiovisual and Digital Communication (ARCOM) is in charge (among other tasks) of collecting content reports from channels to analyze and publish reports of women representation on TV and Radio \cite{arcom24}. 
    These reports are conducted on two \textit{neutral} months which are determined beforehand, to exclude special events such as elections. %
    For the 2023 edition, ARCOM and channels mutually chose May and October. During these months, 41 channels reported a total of 29,707 programs, categorized by program type (e.g. \textit{Information/News}, \textit{Documentary}, \textit{Magazine}, or \textit{Entertainment}).
    In collaboration with ARCOM, the French National Audiovisual Institute (INA) computed the speaking times using an automatic gender classification tool. In 2023, women accounted for only 34\% of the total speaking time, which is significantly lower than the proportion of women in France (51.6\%\footnote{\scriptsize\url{https://www.insee.fr/fr/statistiques/2381474}}).%

    However, few studies have tried to automatically estimate topic disparities based on the gender of speakers.
    In this paper, we want to determine, on a large-scale analysis, if men or women dominate the speaking time on specific subjects.
    We focus on radio and TV news broadcast, but similar analyses should also be carried out for other types of broadcast content, including fiction, documentaries, entertainment programs, etc.
    The contributions of this paper are as follows:
    \begin{enumerate}
        \item We describe, annotate and release a topic-classification dataset of broadcast news extracts ;
        \item We evaluate three different kind of classifiers: baseline BERT models, a few-shot prompted LLM, and Teacher/Student models trained on automatically generated annotations ;
        \item Using the best performing model, we process 11.7k hours of broadcast news to estimate gender representation biases in French audiovisual media, depending on content topics.
    \end{enumerate}

\section{Related works}
    Over the years, automatic topic classification of news has been tackled with different techniques.
    Some methods involve hand-crafted features, keywords counting, Bayesian modeling, or neural networks \cite{doumit2011online,minaee2021deep,cage2020social}.
    More recently, topic classification has been addressed by many using \textit{finetuned} language models based on the Transformer's architecture \cite{grootendorst2022bertopic,flaubertoral,declercq2020_news_topic_classification}. %
    Even Large Language Models (LLM) have been used in conversation mode to categorize news contents \cite{fatemi2023_effectiveness_GPT_IPTC}. These models are often used in a few-shot configuration by prompting the model with the desired task and taxonomy.
    Most of these techniques have been applied for classification of written text news, however, some have been applied on automatic transcripts of audiovisual pre-segmented news subjects \cite{flaubertoral,leopold2007content}. %

    Despite the generalization capabilities praised by some, LLMs face a significant limitation due to their computational power requirements, resulting in substantial financial costs \cite{bogdanov2024nuner}.
    Methods known as knowledge distillation are explored by some to project the LLMs capabilities into smaller, energy efficient models \cite{bogdanov2024nuner,NIPS2014_ea8fcd92,hinton2015distilling,yoo-etal-2021-gpt3mix-leveraging,wang-etal-2021-want-reduce,Smith2023_language_model_loop,zhou2024universalner}.
    These techniques typically employ a large model as the \textit{Teacher}, which generates synthetic training data for the smaller \textit{Student} model.
    In some cases, the \textit{Student} model has been found to achieve better results than the \textit{Teacher} model \cite{wang-etal-2021-want-reduce,zhou2024universalner} for tasks such as Named Entity Recognition, Natural Language Generation and Understanding.

    Datasets and taxonomies related to news classification include AG News, Reuters News, or AFP datasets \cite{minaee2021deep,cage2020social}.
    The International Press Telecommunications Council (IPTC) defines a hierarchical taxonomy of news subjects, comprising 17 top-level categories such as \textit{weather} or \textit{politics}.
    While several studies report high accuracy in automatic IPTC categorization based on text press news \cite{cage2020social,declercq2020_news_topic_classification,fatemi2023_effectiveness_GPT_IPTC}, approaches relying on TV news programs are scarce and have been found to be much more challenging, often requiring manual pre-segmentation of news into homogeneous topics \cite{leopold2007content}. 
\section{Data description}
    We choose to conduct our analysis on French broadcast news, following the scope of ARCOM 2023 channels reports. Using available reports, we filtered programs declared on the \textit{Information/News} type, or broadcasted from 24/7 news cycle channels.
    We preprocess this corpus by transcribing it using an ASR model and merging utterance segments into larger chunks for additional context. Next, we manually annotate a subsample of this dataset (804 dialogues with a total duration of 03h44m).
    
    \subsection{Preprocessing}
            Our first preprocessing step is to transcribe the ARCOM-declared programs using the Whisper model \cite{pmlr-v202-radford23a}, specifically, the model released under \textit{whisper-large-v3} name.
            We use the WhisperX \cite{bain2022whisperx} implementation, offering a speedup of 11.8x over the base implementation.
            ASR evaluations were conducted using an internal French 10h TV news dataset broadcasted from 2019 to 2023, which shares similar properties with our current material, resulting in a WER of 10.66\%.

            A total of 11.7k hours of audiovisual data are transcribed on NVIDIA RTX 2080 Ti GPU cards, with an average speed of 228s per hour of speech. 56.3\% of the transcribed hours are from 24/7 news cycle TV channels, with the remaining from other channels, under the ARCOM \textit{Information/News} type.

            Broadcast news feeds are segmented into utterances using segment-level timecodes obtained from WhisperX.
            On average, these segments have a duration of 4 seconds, which would contain too little information to be efficiently classified into topics. A simple heuristic is implemented to group utterance segments into longer units, which will be referred to as \emph{dialogues}.
            We sequentially merge segments that had a temporal gap with other segments of less than 10s, and a total duration of less than 60s. As a result, dialogues have a duration of 17 seconds.
            Our heuristic does not try to ensure that all dialogues contains only one single news subject. This could impact our follow-up analysis, and we therefore tried to strike a balance between too little context and too many unrelated topics inside a single dialogue.

    \subsection{Annotation guidelines}
        We define a set of 18 topic categories inspired from the IPTC taxonomy.
        Due to the special nature of TV and radio extracts, we decide to add the classes \textit{commercial} for adverts, as well as an \textit{other} category for everything that cannot not be classified in the other 17 categories. For instance, dialogues that are essentially composed of greetings fall in this \textit{other} category. After browsing some examples manually, we decided not to include the \textit{human interest} category, as very few dialogues could be labeled as it.
        The list of categories is shown in Table \ref{tab:list_categories} along with their respective proportions on the annotated dataset.
        \begin{table}[ht]
            \centering
            \caption{List of the available topics, proportions in the 804 annotated dialogues, durations, and Krippendorff's alpha ($\alpha$) agreement score.
            Percentages and durations are shown with averaged agreements between annotators.
            }
                \fontsize{8.5}{8}\selectfont
                \begin{tabular}{rrrc}
                \toprule
                \textbf{Topic} & \textbf{\#Dial.} & \textbf{Duration} & \textbf{$\alpha$}\\
                \midrule
                {religion, belief} & 1.0\% & 02m 23s & 0.50 \\
                {science, technology} & 2.3\% & 05m 17s & 0.36 \\
                {education} & 2.8\% & 06m 53s & 0.52 \\
                {disaster, accident} & 2.9\% & 06m 41s & 0.72 \\
                {labour} & 4.1\% & 09m 26s & 0.49 \\
                {weather} & 4.4\% & 10m 05s & 0.87 \\
                {health} & 4.5\% & 10m 33s & 0.68 \\
                {other} & 4.6\% & 07m 42s & 0.43\\
                {environmental issue} & 6.3\% & 15m 06s & 0.69 \\
                {sport} & 6.5\% & 14m 06s & 0.95 \\
                {lifestyle, leisure} & 6.8\% & 14m 43s & 0.57 \\
                {social issue} & 7.7\% & 17m 50s & 0.25 \\
                {economy, business, finance} & 7.7\% & 18m 37s & 0.74 \\
                {commercial} & 9.4\% & 17m 26s & 0.93 \\
                {arts, culture, entertainment} & 9.8\% & 22m 41s & 0.70 \\
                {crime, law, justice} & 14.5\% & 33m 17s & 0.67 \\
                {politics} & 16.2\% & 39m 57s & 0.62 \\
                {unrest, conflicts, war} & 27.2\% & 1h 05m 51s & 0.82 \\ \midrule
                \multicolumn{1}{r}{\textbf{total}} & 100.0\% & 03h 44m 44s & 0.60 \\
                \bottomrule
                \end{tabular}
            \label{tab:list_categories}
        \end{table}

        Annotators are provided with a short description of the category. They are asked to label a dialogue with all relevant categories (multilabel). The order of categories is not considered important.
        While annotators are encouraged to share their hesitations with others, they are instructed to rely on their own judgments in case of disagreement with other annotators.

        Along with the topic categories, we ask annotators to label each dialogue with the scope significance of the news relative to France (local, national, european, and international) ; whether the subject is about the Russo-Ukrainian war (6.3\% of dialogues) and/or the Israel–Hamas war (16.9\%). Lastly, as the dialogues are automatically assembled from speech utterances without any topic segmentation, we ask annotators to indicate when dialogues are perceived as a succession of different subjects (8.2\%).

    \subsection{Annotation campaign}
        A total of 804 dialogues are randomly selected for manual annotation: 402 from 24/7 news cycle channels (France Info TV, CNews, LCI, France 24, BFMTV), along with 402 dialogues declared under the \textit{Information/News} category from 7 remaining channels (RTL, TF1, M6, RMC, France Info Radio, Europe 1, France 2).
        Corpus annotation is realized by 3 male speech analysis or NLP researchers
        (aged 41, 35 and 26) and a female professional specialized in gender representation issues in media (aged 28).
        To mitigate the subjectivity and complexity of the task, each dialogue is annotated by two annotator, in a mix-and-match way, to compute agreements among all annotators, resulting in 402 dialogues annotated per user. Annotators spent around 7 hours for this task.
        For each dialogue, annotators were able to read the transcription of Whisper, along with the audio and video (for TV programs). %

        The average amount of identified topics per annotated dialogues depending on the annotator varies between 2.55 and 2.62. %
        We compute the inter-annotator agreement using Krippendorff's alpha \cite{krippendorff2018content} for the topic categories. 
        Results are shown in Table~\ref{tab:list_categories} (higher is better). The global alpha is 0.60, indicating diverse annotation strategies, which is consistent with the subjectivity and complexity of this annotation task, together with the dialogue segmentation strategy that results in short out-of-context excerpts.
        Some classes are much more ambiguous, like \textit{social issue} with an alpha of only 0.25. This class could have benefited from a clearer redefinition of its description and range. Conversely, classes like \textit{commercial} and \textit{sport} are (almost) unquestionably labeled by the annotators.

            The annotated dataset\footnote{It can be downloaded free of charge for research purposes at \url{https://www.ina.fr/recherche/dataset-project} under the name \texttt{is24\_news\_topic}.} (03h44m) is randomly splitted into \textsc{dev} and \textsc{test} subsets, with 605 dialogues (75.25\%, 02h50m) for the final \textsc{test} set.
            The full ARCOM 2023 dataset is not public as it requires specific research agreements to be used.

\section{Methodology}
    \label{sec:methodology}
    In this section, we explain the methodology employed to construct classifiers that can predict the subject categories defined in the annotated dataset, from transcribed inputs.
    Our models are evaluated in Precision, Recall and F1-Score, both micro and macro averaged. We compare the predictions of the models with the human annotated \textsc{test} set. As each dialogue is annotated by two people, we weight the cost of an error with the average annotator probability of the topic. 
    If a prediction is made for a category labeled by both annotators, it will count as 1 True-Positive, whereas it will count as 0.5 True-Positive and 0.5 False-Positive if it was annotated by only one annotator. The same principle is applied to negative cases.

    \subsection{Baseline BERT classification models}
        \label{sec:bert_classification_models}
        In the past few years, the most common approach to this kind of classification problems has been to \textit{finetune} a language model such as BERT, with an added classification feedforward layer. We employ this technique as our first \textit{baseline} method. We finetune three kind of French BERT-based language models : CamemBERT \cite{camembert}, FlauBERT \cite{flaubert}, and FlauBERT-Oral \cite{flaubertoral}. 
        CamemBERT and FlauBERT are pretrained with the Masked Language Modeling objective on various text sources such as Wikipedia, books, and web crawls.
        FlauBERT-Oral models are pre-trained on automatically transcribed texts of broadcast news. As such, we expect these models to perform better on our task, compared to the original FlauBERT models.
        Multiple model sizes exists: base (110M to 138M parameters) and large (335M to 373M) variants. Some models are pre-trained on text with case information, while others are pre-trained on uncased text.
        Models are finetuned on the \textsc{dev} set of our annotated corpus. However, we randomly split this set into a training (80\%) and validation (20\%) subsets, so that we can optimize certain hyperparameters and monitor the performances during training without compromising the final annotated \textsc{test} set. Models are trained up to 100 epochs, with a validation every 10 model weight updates. We keep the best performing checkpoint (F1-score) on the validation set.

    \subsection{Mixtral-8x7B few-shot classification}
        As our annotated dataset is rather small, and considered only for development and evaluation purposes, we employed a few-shot classification scheme as a second classification technique.
        We use the Mixtral-8x7B-v0.1-\textit{instruct} \cite{jiang2024mixtral} language model, in conversational mode. The model is made of 47B parameters, but only uses 13B during inference. As of early 2024, it is one of the best performing open-source instruct language models in French \cite{jiang2024mixtral}.
        We use a 4-bit AWQ quantization \cite{lin2023awq} of this model\footnote{\url{https://hf.co/casperhansen/mixtral-instruct-awq}} for it to fit in our GPU cards.
        We prompt it to generate a JSON list of categories contained in the transcript of the dialogue. The list was provided along with the description of all categories. The outputs were post-processed to remove hallucinations, unwanted explanations and newly created classes. Three sample dialogues were included as examples in the prompt, mainly to instruct the model about the required input/output format, and not about the category usage. Those three examples belongs to three categories out of the 18 available. The full prompt along with the post-processing recipe are released with our source code\footnote{\url{https://github.com/ina-foss/is24_news_topic}}.

        This model was used on the annotated \textsc{test} set to assess its performances, as well as on a random sample of unannotated dialogues.
        Using two NVIDIA A100 40Gb GPU cards, we were able to process 353k unannotated dialogues with an average speed of 0.58 dialogues per second. %

    \subsection{Teacher/Student models}
        We explore the use of the Mixtral-generated annotations as a \textit{finetuning} dataset for classification models. 
        As in section \ref{sec:bert_classification_models}, we \textit{finetune} BERT models to classify the dialogues. We use the 353k dialogues annotated by Mixtral (considered as teacher) as the training set. Concerning the \textsc{dev} dataset, used to monitor the F1-score during training, we use the 199 dialogues from the human-annotated \textsc{dev} dataset.
        Unlike in section \ref{sec:bert_classification_models}, the BERT models here (student) are \textit{finetuned} for only 3 epochs, due to the higher quantity of data available for training.
        On a single NVIDIA RTX 4090 GPU with the \textit{camembert-base} (CamB) model, we were able to process around 70 dialogues per second, resulting in a speedup of 120.7x over the Mixtral model which also required GPUs 9.6x more expensive.

        \begin{figure*}[b]
            \centering
            \includegraphics[trim={0 0.25cm 0 0.25cm},clip,width=1\linewidth]{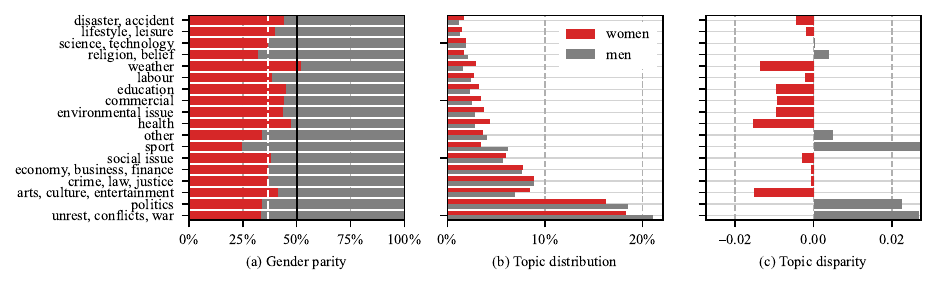}
            \caption{Measured gender representation bias per topic.}
            \label{fig:results_gender}
        \end{figure*}
\section{Results}

    \subsection{Automatic system evaluation}
        \begin{table}[ht]
            \centering
            \caption{F1-score, Precision (P) and Recall (R) on the annotated \textsc{test} set for models: CamB (\textit{camembert-base}), CamL (\textit{camembert-large}), FlauBU (\textit{flaubert-base-uncased}), FlauBC (\textit{flaubert-base-cased}), FlauLC (\textit{flaubert-large-cased}), FlauOF (\textit{flaubert-oral-ft}), FlauOM (\textit{flaubert-oral-mixed}), FlauOA (\textit{flaubert-oral-asr}).}
                \fontsize{8}{8}\selectfont
                \begin{tabular}{crcccccc}
                \toprule
                \multicolumn{2}{r}{\multirow{2}{*}{\textbf{Model}}} & \multicolumn{3}{c}{\textbf{Micro} ($\%$)} & \multicolumn{3}{c}{\textbf{Macro} ($\%$)} \\ \cmidrule(lr){3-5} \cmidrule(lr){6-8}
                &  & \textbf{F1} & \textbf{P} & \textbf{R} & \textbf{F1} & \textbf{P} & \textbf{R} \\
                \midrule
                \multirow{3}[-2]{*}{\textbf{\rotatebox{90}{Baseline}}} & 
                \textbf{CamB} & 50.5 & 77.2 & 37.5 & 24.3 & 82.4 & 20.8 \\
                \textbf{} & \textbf{CamL} & 58.5 & 73.0 & 48.8 & 37.3 & 82.5 & 31.8 \\ \cmidrule{2-8}
                \textbf{} & \textbf{FlauLC} & 55.6 & 69.0 & 46.5 & 40.0 & 69.6 & 34.2 \\ %
                \midrule \midrule
                \multicolumn{2}{r}{\textbf{Mixtral-8x7B}} & 58.6 & 63.0 & 54.8 & 53.8 & 59.8 & 51.5 \\ \midrule \midrule
                \multirow{8}[4]{*}{\textbf{\rotatebox{90}{Teacher/Student}}} & \textbf{CamB} & \textbf{62.5} & 71.7 & 55.3 & \textbf{58.5} & 68.9 & 53.0 \\
                \textbf{} & \textbf{CamL} & 60.9 & 71.7 & 53.0 & 55.9 & 67.3 & 49.8 \\ \cmidrule{2-8}
                \textbf{} & \textbf{FlauBU} & 60.3 & 69.1 & 53.5 & 54.7 & 64.3 & 50.0 \\
                \textbf{} & \textbf{FlauBC} & 60.8 & 70.8 & 53.3 & 56.0 & 66.5 & 49.5 \\
                \textbf{} & \textbf{FlauLC} & 62.2 & 72.9 & 54.3 & 57.3 & 68.7 & 51.4 \\ \cmidrule{2-8}
                \textbf{} & \textbf{FlauOF} & 61.3 & 69.0 & 55.1 & 56.5 & 64.3 & 52.6 \\
                \textbf{} & \textbf{FlauOM} & 62.0 & 74.5 & 53.2 & 54.8 & 67.3 & 48.4 \\
                \textbf{} & \textbf{FlauOA} & \textbf{62.6} & 73.3 & 54.5 & 55.9 & 68.1 & 49.3 \\
                \bottomrule
                \end{tabular}
            \label{tab:results_flaubert}
        \end{table}
        Table \ref{tab:results_flaubert} presents the results obtained with the three types of models detailed in section \ref{sec:methodology}: the finetuned baseline BERT models, the Mixtral-8x7B model in few-shot classification mode, and the Teacher/Student BERT models, finetuned on synthetic annotations generated by the Mixtral-8x7B model.
        Bootstrapping ($N=1000$) confidence intervals \cite{Ferrer_Confidence_Intervals_for} at 95\% are computed. All models obtained an error margin ${\leq} 3.38\%$ for Micro-F1 and ${\leq} 3.96\%$ for Macro-F1.
        For the baseline BERT models, we only show the three best models, namely \textit{camembert-large} (CamL) and \textit{flaubert-large-cased} (FlauLC). 
        While the few-shot prompted LLM model (Mixtral-8x7B) outperforms all baselines models, it is associated with a high computational cost (both inference time and hardware requirements). %
        The Teacher/Student strategy was found to systematically enhance Micro-F1 scores by 2.4 to 15.0 points, and Macro-F1 scores by 17.3 to 34.2 over the baseline models.
        We can notice the imbalanced aspect of the categories, with some of the best models in Micro-F1 having low macro-averaged results.
        Next, we see that training models on synthetic annotations in a Teacher/Student manner allows for better results than the base Teacher model (Mixtral-8x7B). All students models performed better than the base model used to generate the annotations, while having a much lower inference cost.
        It seems the pre-training of FlauBERT-Oral with speech transcripts of broadcast news allows for better results over the FlauBERT base model. However, FlauBERT large model, probably advantaged by its heavier architecture, obtains similar results with Oral models.
        The overall best results are obtained by the \textit{camembert-base} model with a Macro-F1 of 58.5\%, and a Micro-F1 equivalent to the \textit{flaubert-oral-asr} (62.5\% vs 62.6\%). We therefore choose to process the whole 2.1M dialogues dataset using the \textit{camembert-base} model.
        One could argue that the model used to analyze more than 10k hours of speech only obtains a 62.5\% Micro-Averaged F1-score with human generated labels. While true, it is however important to keep in mind that 1) the task in itself is difficult, even for human annotators who only agree with an alpha of $0.60$ ; 2) the metric used is challenging, as annotator disagreements will sanction the model whatever its predictions are ; 3) the end goal is to monitor thematic representation differences between men and women. We can conjecture the model should behave the same way on men and women transcripts. As a result, observed differences between genders should mean there is a bias in the representation of each gender depending on the subject.

    \subsection{Gender representation biases in broadcast news}

        Dialogue-level categories are mapped back to the segment-level speech utterances. We use \texttt{inaSpeechSegmenter} v0.7.7 with its default gender prediction model \cite{doukhan2018open} to measure male and female speech duration of the corresponding segments.
        A recent evaluation of the tool by \cite{inagvad} on a representative dataset of TV news showed an Identification Error Rate (IER) of 6.5\%.
        Globally, women have an average speaking time of 36.58\% (3,417h), out of 9,340 hours of speech predicted as either male or female. 
        Figure \ref{fig:results_gender}.a illustrates instances where women have a lower than average speaking time, in particular in the \textit{sport} category. Conversely, 
        we can see a higher than average speaking time for women in some topics such as \textit{weather}, where the speaking times parity is at 52.0\%, making it the only subject where women are more involved than men. Other higher-than-average topics include \textit{health} (47.4\%), \textit{education} (45.1\%), \textit{disaster-accident} (44.3\%) and \textit{commercial} (44.3\%).
        
        In Figure \ref{fig:results_gender}.b, we notice that the topic distribution is roughly similar to our manually annotated dataset (Table \ref{tab:list_categories}). The topic distribution here is plotted by gender, meaning that at equal speaking time, men are more likely to speak of armed issues (\textit{unrest-conflicts-war}) than women. On the contrary, women are more likely to speak of \textit{arts-culture-entertainment} than men.
    
        Figure \ref{fig:results_gender}.c plots the gender disparity, i.e. the difference between men and women topic usage relative to the global gender parity (36.58\%). 
        This figure allows for better visualization of gender differences present in Figure \ref{fig:results_gender}.b.
        A value greater than 0 means the subject is more predominantly used by men, whereas a value lower than 0 means the subject is more women-specific. It is important to remember that, except for \textit{weather}, none of the topics achieve the gender equality.

        We compute our analysis on both private and public service channels. On 9 public channels (3,678h), women have a speaking time of 40.5\%, while on private channels (5,663h), women speaking time is at 34.1\%. More work is required to describe the factors explaining these differences such as channel editorial, funding and human resources policies. 
        Moreover, we show some news topics suffer from gender biases and if their distribution is not equal across channels, it may impact the speaking time differences between genders. %

\section{Conclusion}
    We focus on gender bias topic representations in French broadcast news programs. We build a corpus of topic classification in order to measure thematic differences between men and women speech. We show that certain subjects remains highly monopolized by men, such as \textit{sports}.
    These results are coherent with previous similar studies, such as the CAC (\textit{Consell de l'Audiovisual de Catalunya}) \cite{ComasdArgemir_2009}, NWCI (National Women's Council of Ireland) \cite{Walsh_Suiter_OConnor_2015} and ARCOM manual thematic studies \cite{arcom24}.
    For future works, we want to explore the other annotations provided in our dataset, in particular the scope significance. %
    The use of the annotated mixed-subjects information along with more advanced methods of topic segmentation into \textit{dialogues} \cite{purver_topic_segmentation} could provide valuable insights, although processing more than 11k hours of speech would require more computational time.
    We also want to conduct an in-depth analysis of representations across public/private and radio/TV channel.
    
    A known limitation of our analysis arises from the stereotypical binary gender categorization, without considering non-binary gender identities. To the best of our knowledge, no automatic tools currently exist for non-binary gender estimation, highlighting the importance of addressing this gap \cite{Ellis2023_nonbinary}. %

\clearpage
\section{Acknowledgements}
This work has been partially funded by the French National Research Agency under the Gender Equality Monitor (ANR-19-CE38-0012) and Pantagruel (ANR-23-IAS1-0001) projects.

\bibliographystyle{IEEEtran}
\bibliography{biblio}

\end{document}